\documentclass[review]{elsarticle}
\PassOptionsToPackage{square,comma,numbers,sort&compress,super}{natbib}
\usepackage{lineno,hyperref}
\usepackage{makecell}
\usepackage{tabularx}
\usepackage{amsmath}
\usepackage{amsthm}
\usepackage{amsfonts}
\usepackage{booktabs}
\usepackage{tabularx}
\usepackage{algorithm}
\usepackage{algorithmic}
\usepackage{multirow}
\usepackage{subfigure}
\usepackage{makecell}
\usepackage{caption}
\usepackage{float}
\usepackage{ulem}
\usepackage{CJKutf8}
\usepackage{bm}

\usepackage{graphicx}
\usepackage{amssymb}
\usepackage{adjustbox}
\usepackage{bm}
\usepackage{chngpage}
\usepackage{array}
\usepackage{colortbl}
\usepackage{threeparttable}
\usepackage{pifont}
\usepackage[dvipsnames]{xcolor}
\definecolor{mygray}{gray}{.9}
\newcommand{\cmark}{\ding{51}}%
\newcommand{\xmark}{\ding{55}}%

\usepackage[numbers,sort&compress]{natbib}

\biboptions{sort&compress}

\journal{Knowledge-Based Systems}
\usepackage{setspace}
\usepackage[numbers,sort&compress]{natbib}
\begin{document}

\begin{frontmatter}

\title{A Knowledge Enhanced Learning and Semantic Composition Model for Multi-Claim Fact Checking}
\author{Shuai Wang}
 \author{Penghui Wei}
 \author[]{Qingchao Kong\corref{mycorrespondingauthor}}
  \cortext[mycorrespondingauthor]{Corresponding author}
 \ead{qingchao.kong@ia.ac.cn}
\author{Wenji Mao}

 \address{State Key Laboratory of Management and Control for Complex Systems, Institute of Automation, Chinese Academy of Sciences, Beijing 100190, China}
\address{School of Artificial Intelligence, University of Chinese Academy of Sciences, Beijing 100049, China}

\begin{abstract}
To inhibit the spread of rumorous information and its severe consequences, traditional fact checking aims at retrieving relevant evidence to verify the veracity of a given claim. Fact checking methods typically use knowledge graphs (KGs) as external repositories and develop reasoning mechanism to retrieve evidence for verifying the triple claim. 
However, existing methods only focus on verifying a \textit{single claim}. As real-world rumorous information is more complex and a textual statement is often composed of multiple clauses (i.e. represented as \textit{multiple claims} instead of a single one), multi-claim fact checking is not only necessary but more important for practical applications.
Although previous methods for verifying a single triple can be applied repeatedly to verify multiple triples one by one, they ignore the contextual information implied in a multi-claim statement and could not learn the rich semantic information in the statement as a whole.
  In this paper, we propose an end-to-end knowledge enhanced learning and verification method for \textit{multi-claim fact checking}. Our method consists of two modules, KG-based learning enhancement and multi-claim semantic composition. 
  To fully utilize the contextual information implied in multiple claims, the KG-based learning enhancement module learns the dynamic context-specific representations via selectively aggregating relevant attributes of  entities.
To capture the compositional semantics of multiple triples, the multi-claim semantic composition module constructs the graph structure to model claim-level interactions, and integrates global and salient local semantics with multi-head attention. 
We conduct experimental studies to validate our proposed method, and experimental results on a real-world dataset and two benchmark datasets show the effectiveness of our method for multi-claim fact checking over KG.

\end{abstract}

\end{frontmatter}

\section{\textbf{Introduction}}

The continuous development of the Internet and social media platforms enables every individual to be a publisher, communicating true or false information instantly and globally.
Among the false  information on the Web, knowledge-based rumorous information accounts for a high proportion, according to the report from an authority website~\footnote{www.xinhuanet.com}\footnote{www.chinanews.com}, which causes severe consequences to individuals and society.
False knowledge misleads the public and undermines their trust in science~\cite{hopf2019fake}.
To inhibit the spread of rumorous information and its serious consequences,  fact checking technique, which aims at retrieving relevant evidence to verify the veracity of given claim(s), is in urgent need.

To verify the veracity of knowledge-based rumorous information, fact checking methods typically use knowledge graphs (KGs) as external repositories and develop reasoning mechanism to retrieve evidence for fact verification. These methods first convert the textual claim to the triple form, and then develop KG-based reasoning mechanism to verify the triple claim. 
Based on the reasoning techniques, they can be classified into path-based and embedding-based methods. Path-based methods verify the triple by reasoning whether there exists a specific link between head and tail entities ~\cite{ciampaglia2015computational,shiralkar2017finding,shi2016discriminative,ijcai2018-522,lin2019discovering,Gad-Elrab2019}, 
while embedding-based methods project KG components into a continuous vector space and calculate a semantic matching score to verify the triple  ~\cite{ammar2019iswc,Dong2019TEKE,Padia2019JWS,nguyen-etal-2020-relational}.

However, existing KG-based fact checking methods only focus on verifying a \textit{single claim} (i.e., a single-claim statement). As real-world rumorous information is more complex and a textual statement is often composed of multiple clauses (i.e. represented as \textit{multiple claims} instead of a single one), multi-claim fact checking is not only necessary but more important for practical applications. Table 1 illustrates an example of multi-claim statement and the corresponding triple claims. 
Although traditional methods for verifying a single triple can be applied repeatedly to verify multiple triples one by one~\cite{pan2018content}, they ignore the contextual information implied in a multi-claim statement and the simple treatment oriented to individual claim could not fully capture the semantic information in the statement as a whole.

\begin{table}
  \caption{A simplified example of \textit{false} statement containing multiple claims.}
\centering
  \label{tab:freq}
  \begin{tabular}{m{5.7cm}|m{5.3cm}}
    \toprule
\hspace{2cm}\textbf{Statement}& \hspace{1cm}\textbf{Triple claim}\\
\midrule
Coffee contains caffeine. Studies show that caffeine can increase tension and accelerate heart rate. Studies also indicate that coffee contains acrylamide,  which can induce cancer.& 1. (coffee, contain, caffeine)\newline 2. (caffeine, increase, tension)\newline 3. (caffeine, accelerate, heart\_rate)\newline 4. (coffee, contain, acrylamide)\newline 5. (acrylamide, induce, cancer)\\
  \bottomrule
\end{tabular}
\label{introexample}
\end{table}

Compared to traditional fact checking technique, multi-claim fact checking poses unique research issues and challenges. First, as the verification result for a multi-claim statement is not simply the conjunction of those of each individual claim. Take Table 1 as an example, if judged in isolation each individual claim is true, but the statement as a whole is verified false, due to the false information “coffee can induce cancer” it implicitly conveys. Thus for multi-claim fact checking, one key research challenge is to represent the compositional semantics of the statement as a whole for fact verification in complex multi-claim situation. Second, multi-claim situation provides additional information to learn context-specific representations of KG components for better verification of the triple claims. Also in Table 1, although the fifth claim “acrylamide can induce cancer” is verified true in single-claim case, if combined with its contextual information in this multi-claim statement, the actual claim should be “acrylamide contained in coffee can induce cancer”. 
Actually the acrylamide contained in coffee is too trivial to induce cancer.
Thus another key research challenge is to enhance entity representation learning with the contextual information in the statement for multi-claim fact verification.

To tackle the above challenges, as entities and relations expressed by other triples provide rich contextual information in the multi-claim statement, this information can be utilized to direct the attention to particular aspects in learning better context-specific representation for the current triple.  Further, for composing the semantic representation of multiple claims, it is important to model claim-level interactions globally so as to acquire the implied semantic relationships among claims in a statement. In addition, composing semantics represented by salient individual claims locally is also important for better verification.

Based on the above considerations, in this paper, we propose a KG-based Learning Enhancement and Semantic Composition method (LESC) for multi-claim fact checking. Our method consists of two modules, KG-based learning enhancement and multi-claim semantic composition. The \textit{KG-based learning enhancement} module learns the dynamic context-specific representations of entities by selectively encoding relevant attributes based on the contextual triple claims in the current statement. To model the inter-claim semantic interactions, Graph neural network (GNN) is a good fit. Previously, GNN was typically used to model the semantic interactions in tasks such as text classification~\cite{kipf2016semi,graphSAGE,velickovic2018graph,graph2018CNN,Yao_Mao_Luo_2019,Sentiment15analysis}, 
syntactic parsing~\cite{ROLE2017labling}, machine translation~\cite{nmt2017syntactic, nmt2018syntactic} and relation extraction~\cite{dt2018re}, with the aim of learning powerful representations for graph nodes. Unlike previous GNN-based methods, we adapt GNN to compose the claim-level semantics for fact verification. The \textit{multi-claim semantic composition} module captures the global and important local semantics implied in multiple claims via devising a customized graph convolutional network, and adopts the multi-head attention mechanism with the HSIC criterion~\cite{hsic05} to compose the local semantics diversely. Finally, the two modules are integrated into a unified framework to train the model in an end-to-end fashion.

Our work has made several contributions:
\begin{enumerate}

\item We address the problem of multi-claim fact checking and propose the first computational model to tackle this problem.

\item We propose a KG-based learning enhancement method to learn context-specific representations of entities by selectively aggregating neighboring attributes based on the contextual information.

\item We propose a graph-based semantic composition method to effectively compose global and local semantics by devising 
a claim-level graph with multi-head attention mechanism.

\item We construct a real-world multi-claim fact checking dataset to verify our method. 
Experimental results on two benchmark datasets and the constructed dataset show the effectiveness of our method for multi-claim fact checking.

\end{enumerate}

The rest of the paper is organized as follows. 
Section 2 introduces the related work on fact checking and KG reasoning.
Section 3 describes in detail our proposed learning enhancement and semantic composition method for multi-claim fact checking. 
In Section 4, we conduct intensive empirical studies to evaluate our work and analyze the experimental results. 
Finally, Section 5 concludes the paper.

\section{\textbf{Related Work}}

In this section, we first review the related work on fact checking, focusing on the KG-based fact checking.
Although the main purpose of KG reasoning is for knowledge graph completion, some KG reasoning methods involve the triple classification task, thus we also review the classification in Section~\ref{kbfc}.  
Most of the KG reasoning methods for link prediction can be adjusted to adapt to the this task,  so that we review the related search in Section~\ref{lp}.

%

\subsection{\textbf{KG-Based Fact Checking}} \label{kbfc}
For fact checking, there are two kinds of external sources to collect evidence from. A class of methods search evidence from  Web texts such as WikiPedia and then develops natural language reasoning methods to verify the given statement~\cite{thorne-etal-2018-fever,hanselowski-etal-2018-ukp,zhou-etal-2019-gear}. 
The other common methods use knowledge bases as external sources, typically using KGs, which contain abundant and high quality facts.

Existing KG-based fact checking methods aim at retrieving evidence from the knowledge graph to verify a claim in the triple form, which can be roughly divided into the following two main groups, including path-based methods and embedding-based methods.

Path-based methods aim at mining the paths between head and tail entities to verify whether there is the specific link between them.
Ciampaglia et al.~\cite{ciampaglia2015computational} proposed the first KG-based fact checking method, which utilized the feature of a single shortest path between head and tail entities to gauge the correctness of a given claim. 
To mine discriminative paths, Shi et al.~\cite{shi2016discriminative} defined some mining rules to leverage different types of information in KG that incorporates connectivity, entity category and predicate interactions. 
The above two methods extract features from relation paths, and then 
verify the truthfulness using a supervised learning framework.
Shiralkar et al.~\cite{shiralkar2017finding} proposed an unsupervised approach that used a flow network to model the multiple paths.
Fionada et al.~\cite{ijcai2018-522} defined some evidence patterns with various optimization techniques to mine evidence.
For effectively finding evidence, Lin et al.~\cite{lin2019discovering} developed a supervised pattern discovery algorithm using ontological information of KG. 
To generate human-comprehensible explanations for candidate facts, Gad-Elrab et al.~\cite{Gad-Elrab2019} defined hand-crafted rules for finding semantically related evidence. 
However, for these path-based methods, effective paths do not always exist due to the incompleteness of real-world KGs.

Embedding-based methods project entities and relations into vector space that can alleviate the above issue. 
Ammar et al.~\cite{ammar2019iswc} adopted RDF2VEC to produce embeddings for triples, and then used RandomForest to classify them, which is a pipeline approach.
TEKE~\cite{Dong2019TEKE} evaluated a triple by measuring the distance between head and tail entities under a specific relation, while it is crispy and cannot adapt to KGs with many kinds of relations.
To learn more robust KG embeddings, Padia et al.~\cite{Padia2019JWS} proposed a linear tensor factorization algorithm to support verification.
To further improve the performance of predicting new triples,
Nguyen et al.~\cite{nguyen-etal-2020-relational} adopted a memory network to encode the potential dependencies among relations and entities, which achieved SOTA results.

However, all the previous methods focus on single-claim fact checking (i.e., the unverified statement is represented by a single triple), ignoring the multi-claim fact checking that is more common in real-world scenarios.
Dual TransE~\cite{pan2018content} is the only method considering this case,
which predicts a truth score for each claim individually with TransE, and then uses the average or minimum of these scores as the final score.
However, Dual TransE verifies each triple claim individually and ignores the interactions among them.
Different from Dual TransE, we model the relationships among the claims using a customized GCN. Experimental results show that our method significantly outperforms Dual TransE.

\subsection{\textbf{Link Prediction on KG}} \label{lp}
Link prediction is the main task in KG reasoning, which aims at knowledge graph completion by ranking the candidates.
Link prediction methods fall into two categories: path-based methods and embedding-based methods.

Path-based methods utilize the relational paths in KG for reasoning.
PRA~\cite{lao2011PRA,LAO2010PRA} employed random walk and path ranking algorithm to extract features from paths between head and tail entities, and then utilized the extracted path features to predict a score with logistic regression.
To select discriminative paths, DeepPath~\cite{das2017deeppath} and MINERVA~\cite{das2018go} utilized reinforcement learning (RL) for effective reasoning. 
The difference between them is that DeepPath needs to give a tail entity while  MINERVA does not, so that MINERVA can make inferences in more difficult cases.
On the basis of MINERVA, Lin et al.~\cite{lin2018multi} proposed Multi-Hop, which adopted a pretrained embedding model to improve the reward quality and proposed a masking mechanism to encourage the model to explore path diversely. 
To alleviate the affect of few-shot relation that cannot provide sufficient triples for robust learning, Lv et al.~\cite{lv-etal-2019-adapting} adopted meta-learning to learn effective meta parameters to solve this issue.
In the above methods, Multi-Hop achieved SOTA results among path-based methods.

Classic embedding-based methods project entities and relations into continuous vector space, including translational distance models and semantic matching models.
TransE is the most representative translational distance model, which wants $h+r\approx t$ when $(h,r,t)$ holds.
However, TransE~\cite{transE} has flaws in dealing with 1-to-N, N-to-1 and N-to-N relations. To this end, some extensions of TransE were proposed to improve the embeddings, including TransH~\cite{transh}, TransR~\cite{transr}, TransD~\cite{transd} and TransM~\cite{transm}.
RESCAL~\cite{rescal} is a representative semantic matching model that uses vectors to represent entities and uses matrices to model the pairwise interactions between entities.
Its extensions includes DistMult~\cite{distmult}, HolE~\cite{HolE2016}, ComplEx~\cite{Complex2016} and ANALOGY~\cite{analog2017} for efficient representation learning. 
Other semantic matching model uses shallow neural network, including NTN~\cite{ntn}, SME~\cite{sme2014} and MLP~\cite{MLP2014}.
In the above methods, DistMult is a simple and efficient one, which can capture the compositional semantics by multiplying the relational matrix. 

Recent methods adopt deep learning to learn better KG embeddings. ConvE~\cite{convE} and ConvKB~\cite{convKB} use CNN to model the interactions among the relation and two entities in the triple. Schilichtkrull et al.~\cite{schlichtkrull2018modeling} introduced relational graph convolutional networks to deal with the highly multi-relational data in KGs. 
A2N~\cite{bansal2019a2n} learns dynamic relation-relevant embeddings with attention mechanism to better predict the missing entity, which achieves the SOTA results on the link prediction task.

Unlike previous fact checking methods that only focus on verifying a single claim, we address the problem of multi-claim fact checking in this paper and propose the first computational model to tackle it. 
To develop effective end-to-end method for multi-claim fact checking, 
we take advantage of the contextual information implied in the multi-claim statement for enhancing entity representation learning. We also represent the compositional semantics of the statement as a whole by modeling the inter-claim interactions for the verification task.

\section{\textbf{Method}}

We define the problem of multi-claim fact checking as follows. 
Given a statement $c$ consists of multiple triple claims $\{(h_{1}, r_{1}, t_{1}), $ $(h_{2}, r_{2}, t_{2}),$ $ ...,(h_{N},$ $r_{N}, t_{N})\}$, where $h_{i},r_{i},t_{i}$ denote the head entity, relation and tail entity of the $i$-th  triple respectively,
the goal of a fact-checking model $f(\cdot)$ is to verify the truth value of the multi-claim statement with the help of a corresponding knowledge graph $\mathcal G$.

\begin{figure*}[t] 
\centering
    \includegraphics[width=1\textwidth]{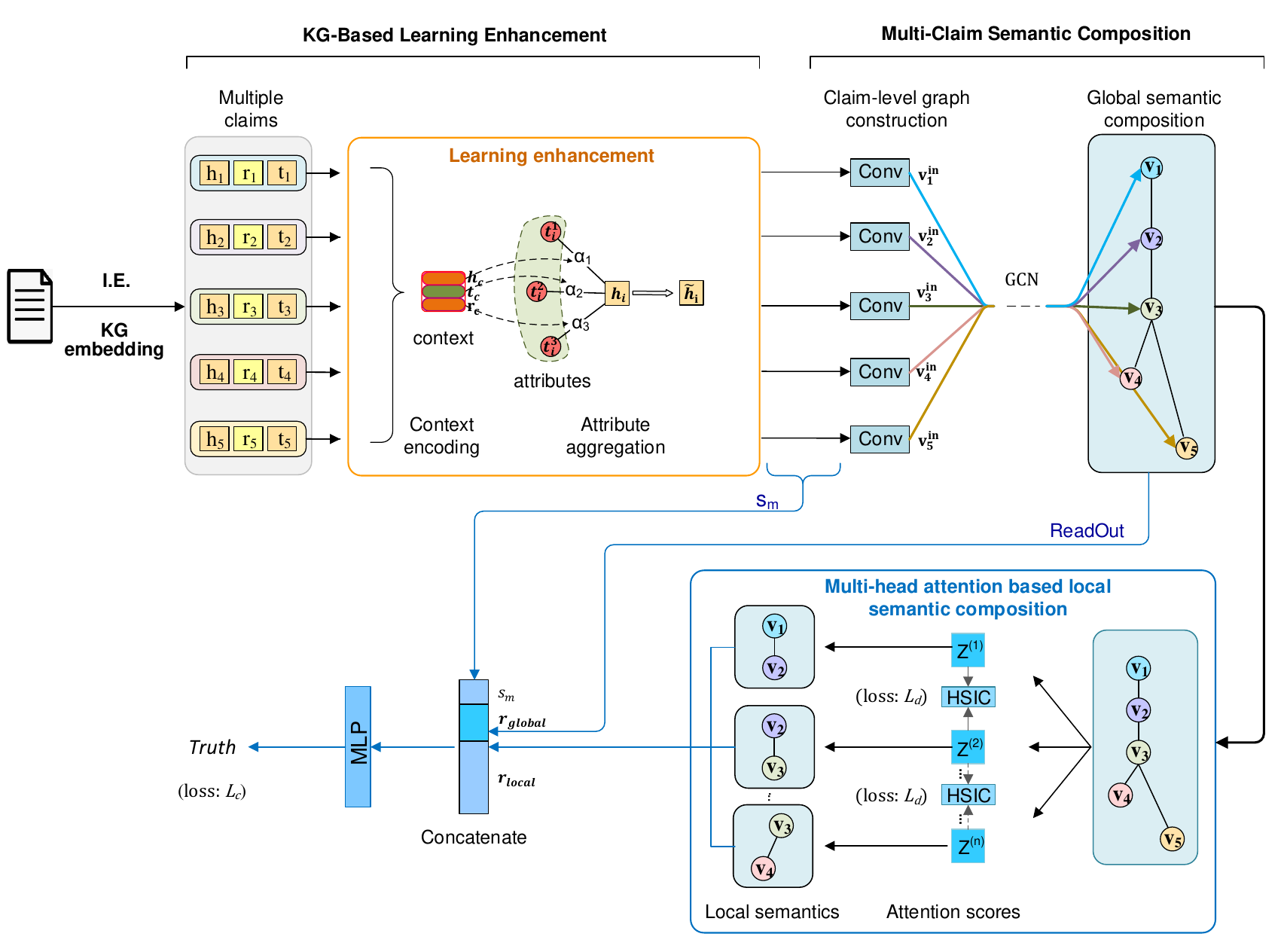}
    \caption{Overall architecture of our KG-based learning enhancement and semantic composition method for multi-claim fact checking.}
\label{mainfig}
\end{figure*}

In this section, we propose an end-to-end \textbf{L}earning \textbf{E}nhancement and \textbf{S}emantic \textbf{C}omposition method (LESC) for multi-claim fact checking
over a knowledge graph.
The overall architecture of LESC is shown in Figure~\ref{mainfig}.
LESC first learns embeddings of entities and relations in the statement,  and then enhances the representations by
a \textit{KG-based learning enhancement} module, which uses the contextual information in the statement to selectively aggregate context-relevant attributes. 
To capture the compositional semantics, \textit{Multi-claim semantic composition} module models the claim-level interactions via developing a multi-head attention GCN, 
and adopts the Hilbert-Schmidt independence criterion for semantic composition.
Finally, LESC learns a final representation for the given statement for verification.

\subsection{\textbf{KG-Based Learning Enhancement}}
To utilize the contextual information in a multi-claim statement, we first encode the representations of the claims within a statement as the context.
Then we use this context representation to selectively aggregate relevant attributes in learning context-specific entity representations.

\subsubsection{\textbf{Encoding contextual information}}

A long statement typically consists of multiple claims, in which other claims in the statement can be viewed as the context of current claim.
Given a statement $c$ with triple claims $\{(h_{1},$$ r_{1},$$ t_{1}),$ $(h_{2}, r_{2}, t_{2}), ...,$ $(h_{N}, $ $r_{N},t_{N})\}$, 
the context representation ($\bm h_c$, $\bm r_c$, $\bm t_c$) is computed by averaging
those of the head entities, relations and tail entities respectively. 
That is, the head context representation is computed as: $\bm h_{c}=\frac{1}{N}\sum_{i=1}^{N}\bm h_i$, and $\bm r_c$ and $\bm t_c$ are computed in the same way.

\subsubsection{\textbf{Aggregating attributes for learning enhancement}}

As the semantic representations of entities can be enriched by their neighboring attributes in KG,
to enhance representation learning of entities in claims, we design an attention mechanism and use the context representation to direct the attention to context-relevant attributes. 
In order to select more relevant attributes, we calculate the attention weights of each attribute based on its semantic distance to the context representation.

Given a KG, each head entitiy $h_i$ has some neighboring attribute entities $\mathcal T_{i}=\{t_{i}^{j}\}_{j=1}^{M}$ with corresponding relations $\mathcal R_{i}=\{r_{i}^{j}\}_{j=1}^{M}$, where $M$ is the number of neighboring attributes of $h_i$.
We take head entities as an example, and the learning enhancement for tail entities is performed similarly.



To compute the attention weights of each attribute, we first adopt a semantic matching function DistMult~\cite{distmult}.
The function is denoted as $f_s(\bm h_c, \bm r_{i}^{j}, \bm t_{i}^{j})$, which gives a higher score to the attribute $\bm t_{i}^{j}$ more relevant to $\bm h_c$. 
We then compute the similarity between neighboring entities $\bm t_{i}^{j}$ and tail context representations $\bm t_c$ using cosine similarity  $f_c(\bm t_c,\bm t_{i}^{j})$. 
We compute $f_c(\bm r_c,\bm r_{i}^{j})$ similarly. 
Finally, we add up these three scores with trainable weights $\omega_{1}, \omega_2, \omega_3$ to obtain the final attention score:
\begin{align}
\begin{split}
\alpha_{i}^{j}&=softmax(\omega_1f_s(\bm h_c, \bm r_{i}^{j}, \bm t_{i}^{j})+\omega_2f_c(\bm r_c, \bm r_{i}^{j})+\omega_3f_c(\bm t_c, \bm t_{i}^{j}))
\end{split}
\label{le}
\end{align}

Based on the above obtained attention scores, the context-specific embeddings of $h_i$ can be calculated as the weighted sum of its neighboring entities. We concatenate it with the initial head embedding and project it to the original dimension to obtain the final embedding $\bm{\tilde{h_i}}$:
\begin{align}
\begin{split}
\hat{\bm h_i}&=\sum_{j=1}^{M}\alpha_{i}^{j} \bm t_{i}^{j}\\
\bm{\tilde{h_i}}&=\bm{W_h}[\bm{h_i}~||~\hat{\bm h_i}]
\end{split}
\label{le}
\end{align}
where $\bm{W_h}$ is the projection matrix, and ``$||$'' denotes the concatenate operation.

In addition, when  the labels of individual claims are available in the training data, they can be used to further enhance the presentation learning.
We adopt the regularized logistic loss to verify each individual triple claim:
\begin{equation}
\mathcal L_t=\sum_{i}log(1+exp(-y_{i}f_s(\bm h_{i},\bm r_{i},\bm t_{i}))),
\label{tv}
\end{equation}
where $y_i \in \{0,1\}$ is the label of the triple claim $(h_{i},r_{i},t_{i})$, and $f_s(\bm h_{i},\bm r_{i},\bm t_{i})$ is the KG scoring function (here we adopt DistMult~\cite{distmult}).

\subsection{\textbf{Multi-Claim Semantic Composition for Verification}}


To better model inter-claim interactions for fact verification, we present each claim as a node and construct a claim-level graph to capture the semantic propagation among claims.
We devise a GCN for global semantic composition.
To capture salient local semantics, we also use a multi-head attention based node selection mechanism to focus on the informative parts in the graph.


\subsubsection{\textbf{Global semantic composition}}

To construct claim-level graph for global semantic composition, we encode each claim using a convolutional layer, and then exploit GCN to model the multi-hop interactions among claims. 
On the basis of this, we use a readout module to acquire the global semantics.



We use $\{v_1, v_2,...,$ $v_N\}$ to represent the multiple claims $\{(h_{1},$$ r_{1},$$ t_{1}),$ $(h_{2}, r_{2}, t_{2}),$ $...,$ $(h_{N}, $ $r_{N},t_{N})\}$.
Similar to ~\cite{convKB,capsule},
we concatenate the representations $[\bm{\tilde{h}}$; $\bm{r}$; $\bm{\tilde{t}}]$ as a $3$-column matrix $\bm U\in\mathbb{R}^{d\times 3}$, and then adopt a convolutional layer with a filter $\bm \omega\in\mathbb{R}^{1\times 3}$ to produce a $d$-dimensional representation $\bm v_i$ for a claim $v_i$.
The process can be formalized as 
\begin{equation}
\bm v(j) ={\rm ReLU}(\bm \omega\cdot \bm{U_{j,:}}+b)
\end{equation}
where $\bm v(j)$ is the $j$-th element of $\bm v$, $\bm U_{j,:}$ denotes the $j$-th row of $\bm U$, b denotes the bias.


To model the interactions among claims, we first add a relation between the claims sharing the same entities to construct the graph structure, and then adopt GCN to capture the semantic propagation.
Specifically, let $\bm V^{in}=({\bm v_1}, {\bm v_2}, ...,{\bm v_{N}})^T$ denote the input representations of claims in the GCN layer, and these claims form a 
graph structure with an adjacent matrix $\bm A$, where $\bm A_{ij}=\bm A_{ji}=1$ if triple $v_i$ shares the same entities with triple $v_j$.
To model the interactions between two-hop connected nodes in the constructed graph, we add a square of the adjacency matrix to create edges between two-hop neighbors.
In the GCN layer, the graph convolution operation in matrix form is defined as:
\begin{align}
\begin{split}
{\bm V^{out}} &= {\rm ReLU}(\bm{\hat{A}}{\bm V^{in}}{\bm W_g^{T}} + {\bm b_{v}})\\
\bm{\hat{A}} &=\bm A + \bm A^{2}		
\end{split}
\label{customizedGraph}
\end{align}
where $\bm W_g^{T}$ and $\bm b_{v}$ are the parameters of the GCN layer.

The output of GCN is the updated claim representations $\bm V^{out}=({\bm v_{1}^{o}}, {\bm v_2^{o}}, ...,$ ${\bm v_{N}^{o}})^T$.
For fact verification, we concatenate the results of mean pooling and max pooling to obtain the global semantic representation ${\bm r_{global}}$:
\begin{equation}
{\bm r_{global}}=\frac{1}{N}\sum_{i=1}^{N}\bm v_i^{o} ~\vert\vert ~{\rm \mathop{max}\limits_{i=1}^{N}}~\bm v_i^{o}
\label{pooling}
\end{equation}
where $||$ denotes the concantenation operation. 
We denote the process expressed in Equation~\ref{pooling} as \textit{ReadOut}.

\subsubsection{\textbf{Salient local semantic composition}}

Since the errors of the statement often occur in a fraction of the claims while most claims in it are correct,
the global representation will flatten the errors in this case.
To focus on more informative local semantics, we adopt self-attention mechanism to compute important scores for each claim based on both node features and graph topology. 
To improve the local semantic composition, we further use multi-head attention to capture multiple local semantics, and then adopt HSIC criterion to enhance the disparity of the attention scores. 
In addition, we also verify the  correctness of each claim.

To focus on salient local semantics, we compute the attention scores $\bm Z \in \mathbb{R}^{N\times 1}$ 
to represent the importance of each node using another graph convolution layer:
\begin{equation}
{\bm Z}={\rm tanh}(D^{\frac{1}{2}}\bm AD^{-\frac{1}{2}}\bm{V}^{out}\Theta_{att}),
\label{localAtt}
\end{equation}
where $D\in \mathbb{R}^{N\times N}$ is the degree matrix of $\bm A$, $\Theta_{att}\in \mathbb{R}^{N\times 1}$ is the parameter of this self-attention layer.
Based on the computed attention scores $\bm Z$, we then select the top $k$ claims :
\begin{align}
\begin{split}
idx={\rm{TopRank}}(\bm Z,k)\\
\bm V^{(l)}=\bm V_{idx}\odot \bm Z_{idx}
\end{split}
\label{select}
\end{align}
where TopRank  returns the index top $k$ values, $idx$ is the index of the selected claims,
$\bm V_{idx}\in \mathbb{R}^{k\times d}$ denotes the selected claims from $\bm V^{out}$, $\bm Z_{idx}\in \mathbb{R}^{k\times 1}$ denotes the corresponding attention scores, $\cdot_{~idx}$ is an indexing operation and $\odot$ is the element-wise multiplication.
We then also use the \textit{ReadOut} method in Equation~\ref{pooling} to get the salient local representations $\bm r_{local}$ from $\bm V^{(l)}$.

In previous studies,  attention-based node selection method is used 
to coarsen the graph for capturing the structural information of graphs~\cite{pmlr-v97-lee19c}. Our work adopts the node selection technique for local semantic composition.

\textbf{Multi-head attention based diverse composition} \quad 
In addition, to focus on multiple parts of the graph, we adopt multi-head attention to produce multiple attention scores $\mathcal Z = \{\bm Z^{(i)}\}_{i=1}^{n_{a}}$, where $n_{a}$ is the number of attention heads. 
Accordingly, we obtain multiple 
local representations $\mathcal R_{local}=\{\bm r_{local}^{i}\}_{i=1}^{n_{a}}$ based on the different attention scores.
We concatenate all the $\bm r_{local}^{i} \in \mathcal R_{local}$  as the final local representation:
${\bm r_{local}}=\bm r_{local}^{1}||\bm r_{local}^{2}||...||\bm r_{local}^{n_{a}}$.

To encourage these multiple attention heads to select nodes diversely via producing different attention scores, we employ the Hilbert-Schmidt Independence Criterion (HSIC)~\cite{hsic05} to enhance the disparity of each two attention scores in $\mathcal Z$. 
Given a pair of attention scores $\{\bm Z^{(a)},\bm Z^{(b)}\}\in \mathcal{Z}$ where $a\not=b$, the HSIC constraint of them is denoted as ${\rm HSIC}(\bm Z^{(a)},\bm Z^{(b)})$. The corresponding loss function can be formalized as
\begin{align}
\mathcal L_{d}&=\sum_{\{\bm Z^{(a)},\bm Z^{(b)}\}\in \mathcal{Z}}{\rm HSIC}(\bm Z^{(a)},\bm Z^{(b)}) \nonumber
\\
 &=\sum_{a,b} (d-1)^{-2}tr(\bm{RK}^{(a)}\bm{RK}^{(b)})
\label{HSIC}
\end{align}
where $\bm K^{(a)},\bm K^{(b)},\bm R\in \mathbb{R}^{d\times d}$, $\bm K^{(a)}_{i,j}=\langle \bm Z^{(a)}_{i}, \bm Z^{(a)}_{j}\rangle$, $\bm K^{(b)}_{i,j}=\langle \bm Z^{(b)}_{i}, \bm Z^{(b)}_{j}\rangle$, where $\langle \cdot,\cdot \rangle$ denotes inner product.
And $\bm{R}=\bm{I}-\frac{1}{d}\bm{ee}^{(T)}$, where $\bm I$ is an identity matrix and $\bm e$ is an all-one vector.

HSIC is a kind of non-parametric independence measure, which has been used for learning robust regression and classification~\cite{pmlr-v119-greenfeld20a,10.1145/3394486.3403177}.
Here we exploit HSIC for enhancing the diversity of semantic composition.

\textbf{Incorporating individual claims} \quad 
For fact verification, it also needs to verify the correctness of each claim.
We also adopt DistMult to predict scalar scores  $[s_{1},s_{2}, ...,s_{N}]$ for each claim, which is formulized as follows:
\begin{equation}
s_{i}=f_s(\bm h_{i},\bm r_{i},\bm t_{i})=\bm{\tilde{h_{i}}^{T}}Diag(\bm{r_{i}})\bm{\tilde{t_{i}}}~.
\end{equation}
where $Diag(\bm r)$ is a diagonal matrix formed by the elements in $\bm{r_{i}}$.

We select the minimum one from all the scores (i.e., the score of the most implausible one) as the representative, which is denoted as $s_{m}=Min([s_{1},s_{2}, ...,s_{N}])$. 
The reason is that the statement is false if any claim in it is false.

\subsection{\textbf{Final Verification and Optimization}}
The model verifies the multiple triples from three aspects via concatenating the scalar score $s_{m}$, global representation $\tilde{\bm v}^{(g)}$ and local representation $\tilde{\bm v}^{(l)}$ to obtain the final represenation.
The representation is fed into a multi-layer perceptron (MLP) with parameters $\bm W_{1}$, $\bm W_{2}$, $\bm b_{fv}$. A sigmoid function $\sigma$ is adopted to predict a final score $s_y\in(0,1)$ for the statement:
\begin{equation}
s_{y} = \sigma ({\bm W_2} {\rm tanh}(\bm W_{1}[s_{m}||\bm r_{global}||\bm r_{local}]+{\bm b_{fv}})).
\label{fv}
\end{equation}

We also adopt the logistic loss $\mathcal L_{c}$ to encourage the model to predict a higher score for a true statement than a false one.
\begin{equation}
\mathcal L_t=\sum_{i}log(1+exp(-y^{i}s_{y}^{i}))),
\label{tv}
\end{equation}
where $s^{i}_{y}$ and $y^{i}\in$\{0,1\} denote the predicted score and label of the $i$-th statement respectively. 
We optimize the following loss function to train our model in an end-to-end fashion:
\begin{equation}
\mathcal L= \mathcal L_{c}+\lambda_{1} \mathcal L_{t}+\lambda_{2} \mathcal L_{d},
\label{finalLoss}
\end{equation}
where $\lambda_{1}$ and $\lambda_{2}$ are the trade-off parameters.

\section{\textbf{Experiments}}
In this section, we evaluate our LESC method by comparing it with the previous fact checking methods and representative KG reasoning methods in the related work. 
We then analyze the experimental results in detail.
\subsection{\textbf{Datasets}} \label{datasets}

We construct the first multi-claim fact checking dataset for our study\footnote{We shall release the source code and datasets.}.
We first construct a real-world KG in food domain (named \textit{FOOD}). We collect various foods and their corresponding effects 
from a well-known food website\footnote{https://www.meishichina.com/}, where food effects are presented as verb phrases in list form.
We treat foods, and verbs and nouns in verb phrases as head entities, relations and tail entities respectively.
We also collect ingredients and efficacies of foods from another popular website~\footnote{http://www.chinanutri.cn/}, and then add the corresponding entities and relations \textit{contain}  (for ingredients) and \textit{effect} (for efficacies) to form triples.

We then construct the multi-claim training data for \textit{FOOD}. 
To be compatible with the realistic statement, we generate samples by random work(RW) on \textit{FOOD}.  
Starting node of RW is a food entity. 
The RW module randomly walks one to four steps at a time, and it is performed one to three times per sampling.
A sample is true if the food contains the corresponding ingredients or effects, otherwise, it is false.
Table~\ref{claimexample} shows some generated examples.
In addition, we adopt the random negative sampling~\cite{transE} to generate negative samples by replacing a correct triple in the claim with a false one. The labels of each individual claim is given in the training process.
For test dataset, we collect food-related statements from a fact-check website\footnote{http://www.piyao.org.cn/}. 
We extracted several claims (i.e., triples) from the textual statement using MinIE~\cite{gashteovski2017minie}.
These extracted real-world claims serve as test data for \textit{FOOD}.

We also conduct experiments on two most commonly used KGs \textit{FB15K} and \textit{FB15K-237} in KG reasoning tasks.  For \textit{FB15K} and \textit{FB15K-237}, we also generate training and test samples by RW with the same strategy. 
The statistics of the three datasets are shown in Table~\ref{data}.


\subsection{\textbf{Experimental Setup}}
\subsubsection{\textbf{Hyperparameters}}
We pretrain 18-dimensional representations of entities and relations by DistMult for two datasets, and they are optimized during the training process. 
To accelerate the convergence of our model, we adopt ConvKB~\cite{convKB} to pretrain CNN for claim-level graph construction. We set $k$ in Equation~\ref{select} to 2, and the number of attention head $n_a$ to 2.
The trade-off parameter $\lambda_1$ and $\lambda_2$ are set to 1 and 0.1 respectively. We train the model with 100 batch size and 0.001 learning rate using AdaGrad. We adopt L2-regularization to avoid overfitting. 

\subsubsection{\textbf{Comparative Methods}}
Most previous methods are designed for single-claim fact checking. Dual TransE~\cite{pan2018content} is the only method 
that has considered to handle the multi-claim fact checking, while it only uses transE to predict a score for each triple individually.
In addition, we choose a classic embedding-based method DistMult~\cite{distmult}, a SOTA KG reasoning method A2N~\cite{bansal2019a2n}, a recent path-based method Multi-Hop~\cite{lin2018multi} as well as a SOTA fact-checking method R\_MeN~\cite{nguyen-etal-2020-relational} for comparison.
\begin{itemize}
\item{DistMult}~\cite{distmult} is a classic semantic matching model, which achieves superior performance by adopting a bilinear model.
\item{A2N}~\cite{bansal2019a2n} dynamically aggregates neighboring entities in KG according to the input triple, which achieves SOTA results in knowledge graph completion task.
\item{Multi-Hop}~\cite{lin2018multi} uses reinforcement learning to sequentially extend the path from the head entity to tail entity, using pretrained KG embedding model and mask mechanism for efficiently reasoning.
\item{R\_MeN}~\cite{nguyen-etal-2020-relational} uses a memory module to encode the potential dependencies among the relations and entities for effectively predicting new facts.
\end{itemize}

Since each of these methods produces a score list for a claim, we obtain the final score of the claim by 
selecting the minimum one as the representative.
Then we classify the given statement by a threshold that is optimized on the validation set. 
We use accuracy and $F_1$ as the evaluation metrics.

\begin{table}
\caption{Examples of multiple triples obtained by random walk}
\centering
\begin{tabular}{c|cc}
    \toprule
\multirow{3}*{True}&1.& tangerine $\xrightarrow{\makebox[12mm]{contain}}$ VC $\xrightarrow{\makebox[12mm]{improve}}$ immunity\\
\cmidrule(r){2-3}
{}&2.&\makecell[c]{henapple $\xrightarrow{\makebox[12mm]{contain}}$ lecithin $\xrightarrow{\makebox[12mm]{contain}}$ \\ $\hookrightarrow$ phosphatidylcholine$\xrightarrow{\makebox[12mm]{maintain}}$ colon}\\
    \midrule
\multirow{2}*{False} &1.& cherry $\xrightarrow{\makebox[12mm]{contain}}$ cyanide $\xrightarrow{\makebox[12mm]{cause}}$ poisoning\\
\cmidrule(r){2-3}
{}&2.&coffee $\xrightarrow{\makebox[12mm]{contain}}$ acrylamide $\xrightarrow{\makebox[12mm]{cause}}$ cancer\\
  \bottomrule
\end{tabular}

\label{claimexample}
\end{table}

\begin{table}[t]
\caption{Statistics of the datasets}
\resizebox{1\textwidth}{!}{
\setlength{\abovecaptionskip}{3pt}%
\setlength{\belowcaptionskip}{0pt}%
  
\centering
  \begin{tabular}{c|ccc|ccccc}
    \toprule
    \multirow{2}*{Dataset}&\multicolumn{3}{c|}{KG}&\multicolumn{5}{c}{Claim}\\
\cmidrule [0.2 pt]{2-9}
{}& \#entity & \# relation &\#triples &\#train &\#valid &\#test&\makecell[c]{\#Avg. of triples\\ per claim} &\makecell[c]{\#Max. of triples\\ per claim} \\
\hline
FOOD & 4192	&86 & 26,767&  51,250&	6,200& 876 & 4.2 & 12\\
FB15K-237& 14,505& 237 & 272,115& 4,575,000 & 258,050& 292,700 & 5.6 & 12\\
FB15K & 14,951	&1,345 & 483,142&  5,285,400& 276,453& 292,100 & 5.6 & 12\\
  \bottomrule
\end{tabular}
}

\label{data}
\end{table}

\begin{table}[t]

\caption{Experimental results on fact checking by different methods.}
\centering
 \renewcommand\tabcolsep{5.0pt}
\begin{threeparttable}
  \begin{tabular}{ccccccc}
  \toprule [1.2 pt]
    \multirow{2}*{\textbf{Method}}  & \multicolumn{2}{c}{\textbf{FB15K}} & \multicolumn{2}{c}{\textbf{FB15K-237}}  & \multicolumn{2}{c}{\textbf{FOOD}}\\
\cmidrule(lr){2-3} \cmidrule(lr){4-5} \cmidrule(lr){6-7} 
	{}&{Acc.}&$F_1$&{Acc.}&$F_1$&{Acc.}&$F_1$\\    
    \hline
Dual TransE       &0.830&0.823&0.813&0.810	 &0.803 & 0.802~\\
DistMult      &0.844&0.837&0.820&0.821	 &0.822&0.820~\\
Multi-Hop  &0.860&0.847&0.830&0.824	 &0.824&0.821~\\
R\_MeN     &0.870&0.866&0.840&0.831	 &0.825&0.822~\\
A2N        &0.873&0.869&0.852&0.846 &0.830&0.824~\\
\hline
LESC (Our)       &\textbf{0.894}&\textbf{0.880}&\textbf{0.873}&\textbf{0.869} &\textbf{0.861}&\textbf{0.860}~\\
\bottomrule [1.2 pt]
\end{tabular}
\end{threeparttable}
  
\label{mainResults}
\end{table}

\subsection{\textbf{Experimental Results}}
\subsubsection{\textbf{Main Results}}
Table~\ref{mainResults} shows the experimental results. It can be seen from the table that our method achieves the best results on both evaluation metrics, and it outperforms other methods by a large margin on three datasets.
R\_Men and Multi-Hop are superior to DistMult and Dual TransE, and A2N is the best model among the comparative methods.
A2N combines the relation-relevant graph neighbors of an entity to learn representations, which is beneficial for two-hop reasoning.
Compared to A2N, our method use contextual information to selectively compose neighboring attribute entity for learning context-specific representations, which are more informative and powerful.

Our method outperforms other methods significantly, especially on the \textit{FOOD} dataset. The reason of this is that our LESC can capture the latent compositional semantics, while other methods cannot.
It also indicates that our method can learn better representations of the statements for fact verification. 
The results demonstrate the effectiveness of our method for multi-claim fact checking.

\subsubsection{\textbf{Ablation Study}}
To verify the effectivenss of each component in our model, we construct six variations of LESC:

\begin{itemize}
\item $-\mathcal{L}_t$: excludes the supervised signal of each single triples in Equation~\ref{tv}.
\item $-\mathcal{L}_d$: excludes HSIC in Equation~\ref{HSIC}.
\item $-$LE: excludes the learning enhancement with contextual information in Equation~\ref{le}.
\item $-$GSL: excludes the global semantic learning in Equation~\ref{pooling}.
\item $-$LSL: excludes the local semantic learning in Equation~\ref{localAtt}--\ref{select}.
\item $-$GSL$,-$LSL: excludes both two modules simultaneously.
\end{itemize}

Table~\ref{ablationStudy} shows the experimental results of the ablation study. The performance slows down without each module. 
``$-\mathcal{L}_t$" demonstrates that the multi-task learning framework can facilitate the fact checking task, because it can provide an additional supervised signal to help the model learn better representations. 
``$-\mathcal{L}_d$'' indicates that  HSIC criterion facilitates the model to learn better representations.
Results of ``$-$LE'' show that the learned context-specific representations are more powerful for learning enhancement.
The performances drop significantly without ``$-$GSL'' or ``$-$LSL'', especially in \textit{FOOD}, which demonstrates the 
importance of modeling the message passing using GCN for verifying multi-claim statement.
The experimental results on the ablation study further verify the usefulness of each component in our method.

\begin{table}[t]
\caption{Experimental results of the ablation study. ``$^\ast$'' denotes that the performance drops significantly while excluding the module.}
\centering
  \begin{tabular}{lcccccc}
  \toprule [1.2 pt]
    \multirow{2}*{\textbf{Method}}  & \multicolumn{2}{c}{\textbf{FB15K}} & \multicolumn{2}{c}{\textbf{FB15K-237}}& \multicolumn{2}{c}{\textbf{FOOD}}\\
\cmidrule(lr){2-3}\cmidrule(lr){4-5} \cmidrule(lr){6-7}
	{}&{Acc.}&$F_1$&{Acc.}&$F_1$&{Acc.}&$F_1$\\    
    \hline
Full Model                 &0.894&0.880&0.873&0.869 &0.861&0.860~\\
\hline
$- \mathcal{L}_t$       &0.889&0.875&0.869&0.865 &0.854&0.853~\\
$- \mathcal{L}_d$      &0.884&0.871&0.867&0.864 &0.846&0.843~\\
$-$LE                         &0.874&0.860&0.858&0.850 &0.836&0.824~\\
$-$GSL                      &0.876&0.863&0.855&0.843 &0.835&0.825~\\
$-$LSL                       &0.871&0.862&0.850&0.837 &0.831$^\ast$&0.822$^\ast$~\\
$-$GSL,$-$LSL         &0.862&0.854&0.844&0.830 &0.824$^\ast$&0.814$^\ast$~\\

\bottomrule [1.2 pt]
\end{tabular}
\label{ablationStudy}
\end{table}

\subsection{\textbf{Further Analysis}}
We conduct the following additional experiments to further demonstrate the effectiveness of our model.
We first analyze the effect of the number of claims in a statement.
We then conduct the parameter analysis experiments to analyze each module in detail.

\subsubsection{\textbf{Effect of the number of claims in a statement}}

\begin{figure}[t] 
\begin{flushleft}

\centering 
\subfigure[FB15K]{ 
\includegraphics[width=0.31\textwidth]{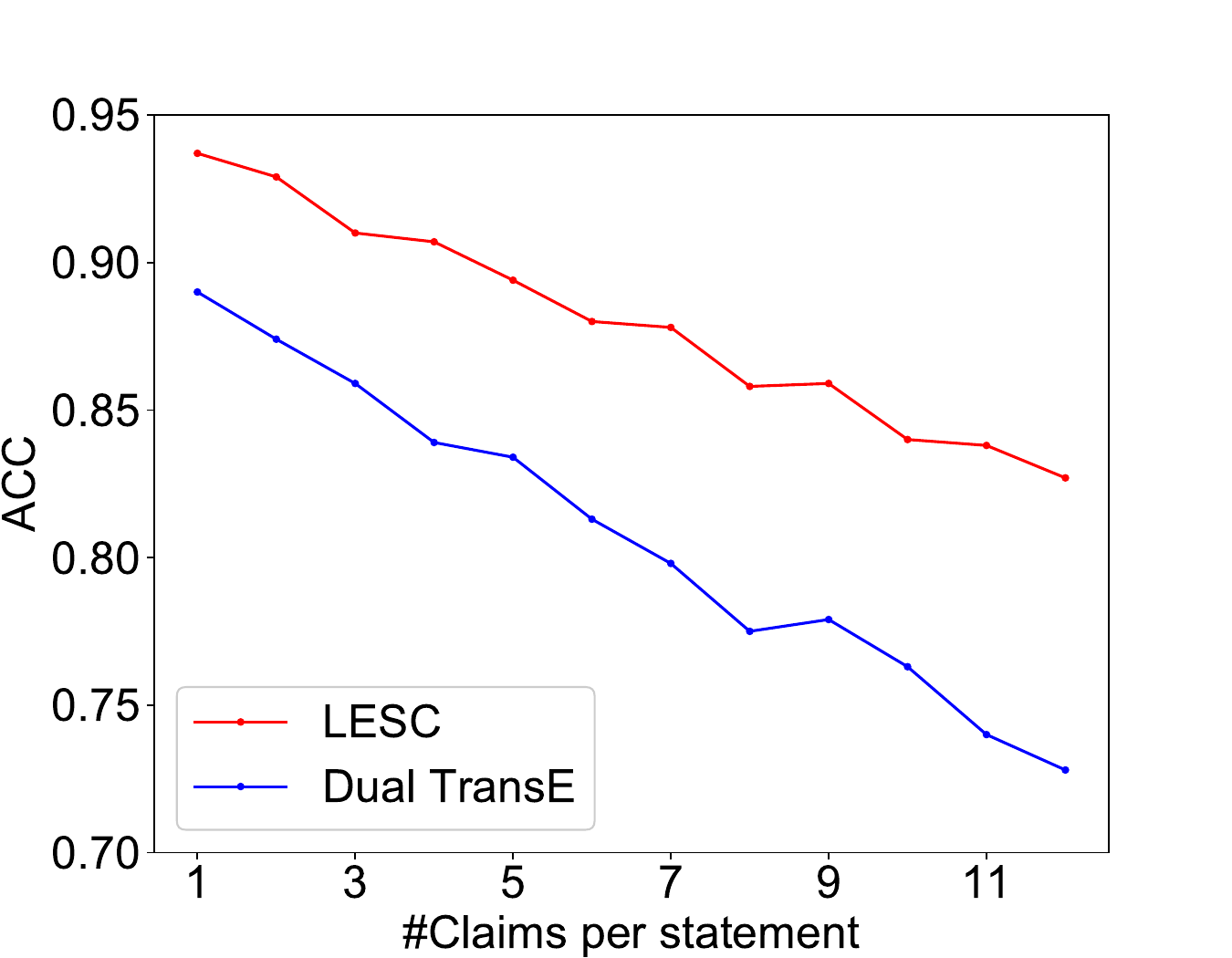}}
\subfigure[FB15K-237]{ 
\includegraphics[width=0.31\textwidth]{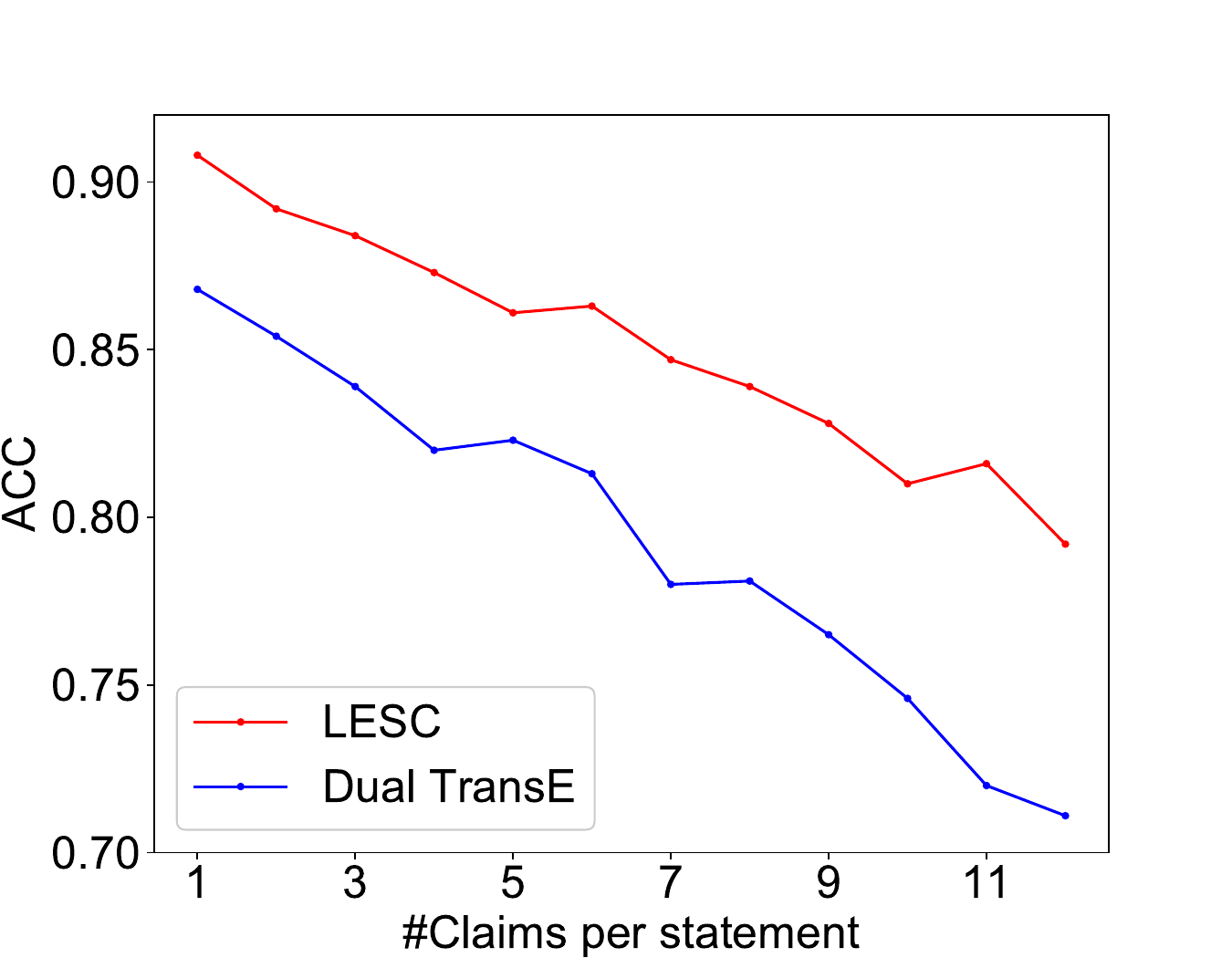}}
\subfigure[FOOD]{ 
\raggedright{\includegraphics[width=0.31\textwidth]{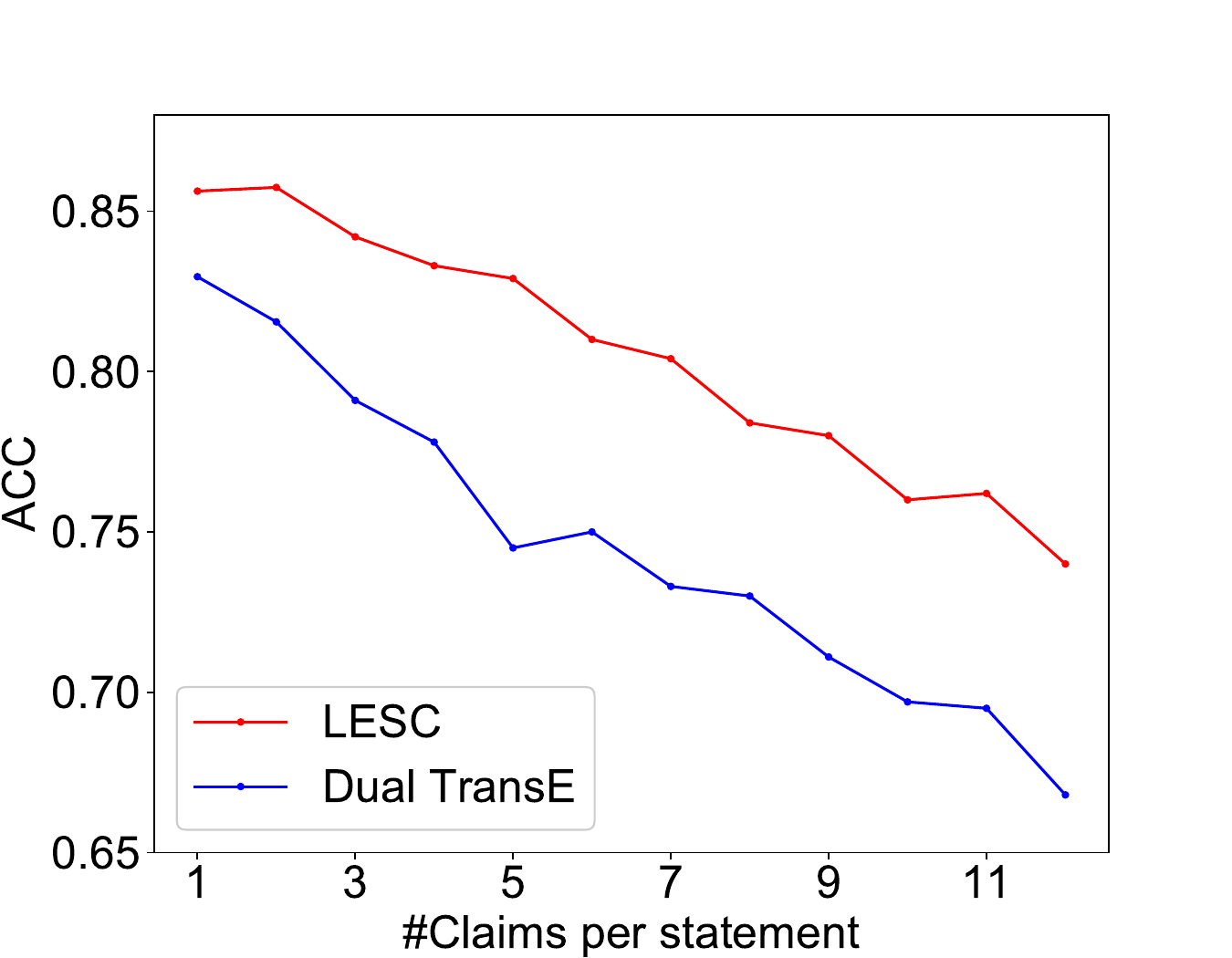}}}

\caption{Comparative results for the statement with different numbers of claims.} 
\label{Fig.lable} 
\end{flushleft}
\end{figure}

To explore the influence of the number of multiple claims, we divide the test sets according to the number of claims that each statement has.
We choose Dual TransE for comparison to show the effectiveness of our model. The experimental results in Figure~\ref{Fig.lable} shows that our method outperforms Dual TransE in every case, especially in the cases that one statement contains more than one claims.
The performance of Dual TransE drops sharply as the number of claims increases.
The reason of this is that if one claim is mispredicted, the final prediction for the claim will be wrong. 
Thus, the more claims contained in a statement, the more error-prone the model is.

It can be seen that our method drops more slowly compared to Dual TransE. The reason is that our method can make better use of contextual information and capture the compositional semantics to learn more powerful representations.
Especially on the \textit{FOOD}, our method even achieves the best performance on 2-claim statements, indicating that modeling the interactive relationship among claims is very conducive to improving performance.
Even for the statements having more than five claims, our model can achieve acceptable results and outperform Dual TransE by a large margin.
The experimental results demonstrate the effectiveness of our method for verifying the statement having multiple claims.

\subsubsection{\textbf{Effect of Customized Graph Convolution}}
To show the effect of our customized graph convolution operation (Equation~\ref{customizedGraph}), we further construct the adjacent matrix of a graph in three different ways for comparison, including \textit{fully connected graph}, ``$\hat{A}=A$'' and ``$\hat{A}=A^2$''.
Table~\ref{graphs} shows the results.
The fully connected graph has the worst performance, because it includes unwanted edges that make the graph very noisy.
Our customized GCN outperform ``$\hat{A}=A$'' and ``$\hat{A}=A^2$'' obviously, especially on \textit{FOOD}.
Since the receptive field of the original GCN ``$\hat{A}=A$'' is restricted, it cannot model multi-hop interactions without stacking multiple layers.
``$\hat{A}=A^2$'' can model the two-hop interactions on the graph for learning more informative representations, so that it performs better. 
Intuitively, the strength of one-hop relations should be stronger than that of two-hop, and our customized GCN can reflect this. The customized GCN achieves the best results among all the variations.

\begin{table}[t]
\caption{Experimental results of constructing the adjacent matrices of graphs in different ways.``\dag'' means that the results outperform the original graph ``$\hat{A}=A$'' by paired t-test at the significance level of 0.01.}
\centering
 \renewcommand\tabcolsep{5.0pt}
\begin{threeparttable}
  \begin{tabular}{ccccccc}
  \toprule [1.2 pt]
    \multirow{2}*{\textbf{Graph}}  & \multicolumn{2}{c}{\textbf{FB15K}} & \multicolumn{2}{c}{\textbf{FB15K-237}}  & \multicolumn{2}{c}{\textbf{FOOD}}\\
\cmidrule(lr){2-3} \cmidrule(lr){4-5} \cmidrule(lr){6-7} 
	{}&{Acc.}&$F_1$&{Acc.}&$F_1$&{Acc.}&$F_1$\\    
    \hline
Fully connected       &0.879&0.868&0.860&0.854	 &0.837 & 0.831~\\
\hline
$\hat{A}=A$             &0.890&0.878&0.865&0.863	 &0.848&0.843~\\
$\hat{A}=A^{2}$       &0.887&0.874&0.868&0.865	 &0.853&0.847~\\
$\hat{A}=A+A^{2}$ (Our)  &\textbf{0.894}&\textbf{0.880}&\textbf{0.873}&\textbf{0.869} &\textbf{0.861}$^\dag$&\textbf{0.860}$^\dag$~\\
\bottomrule [1.2 pt]
\end{tabular}
\end{threeparttable}
\label{graphs}
\end{table}

\subsubsection{\textbf{Parameter Analysis}}

\begin{figure}[t] 
\begin{flushleft}
\centering 
\subfigure[$\lambda_1$]{ 
\raggedright{\includegraphics[width=0.43\textwidth]{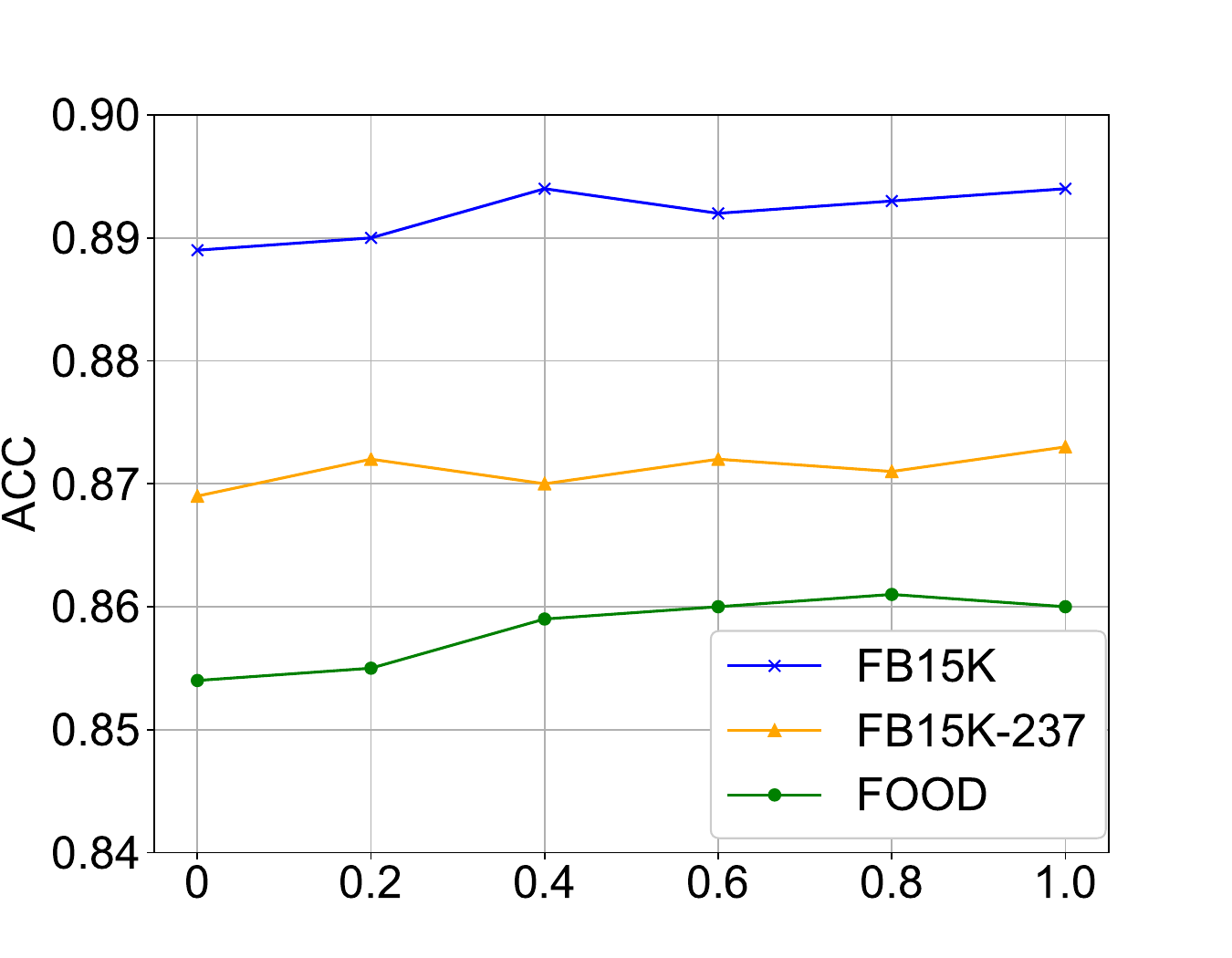}}}
\subfigure[$\lambda_2$]{ 
\includegraphics[width=0.43\textwidth]{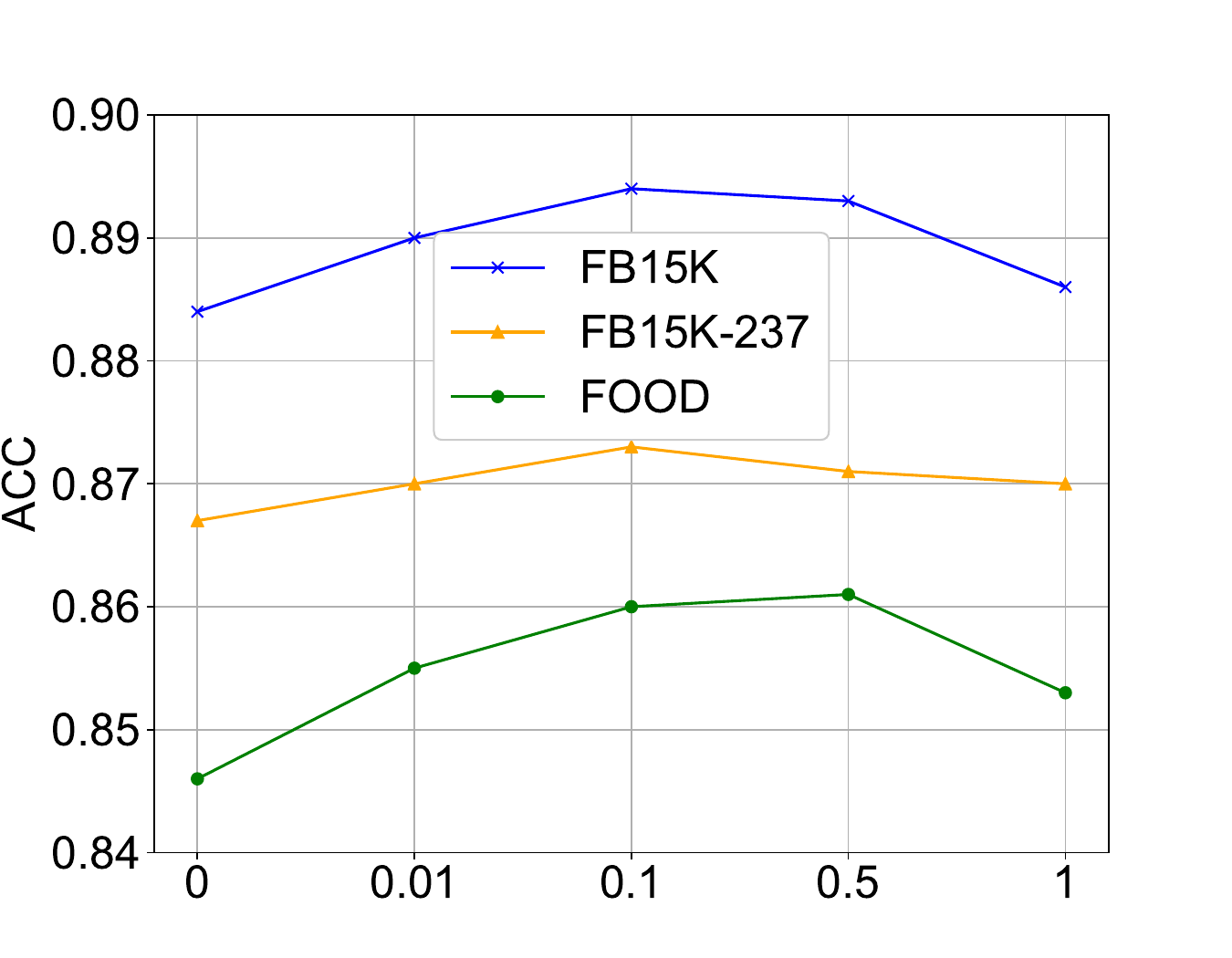}}
\caption{Impact of $\lambda_1$ and $\lambda_2$ in loss function~\ref{finalLoss}} 
\label{parameter} 
\end{flushleft}
\end{figure}

To measure the impacts of fluctuation in parameters of our model, we conduct the following parameter sensitivity experiments.
We first analyze the hyper-parameters in the final loss function (i.e., Equation~\ref{finalLoss}), and Figure~\ref{parameter} shows the results.
We vary the values of $\lambda_1$ from 0 to 1.
The performance becomes better as $\lambda_1$ increases, indicating that our model benefits from the supervised signals of individual claims.
In order to show the influence of $\lambda_2$ clearly, we use the log axis.
With the growth of $\lambda_2$, the accuracy goes up at first and then decrease, achieving the best results at 0.1.
The reason of this is that too large $\lambda_2$ will mislead the model.

\begin{figure}[t] 
\begin{flushleft}
\centering 
\subfigure[\#attention heads]{ 
\raggedright{\includegraphics[width=0.43\textwidth]{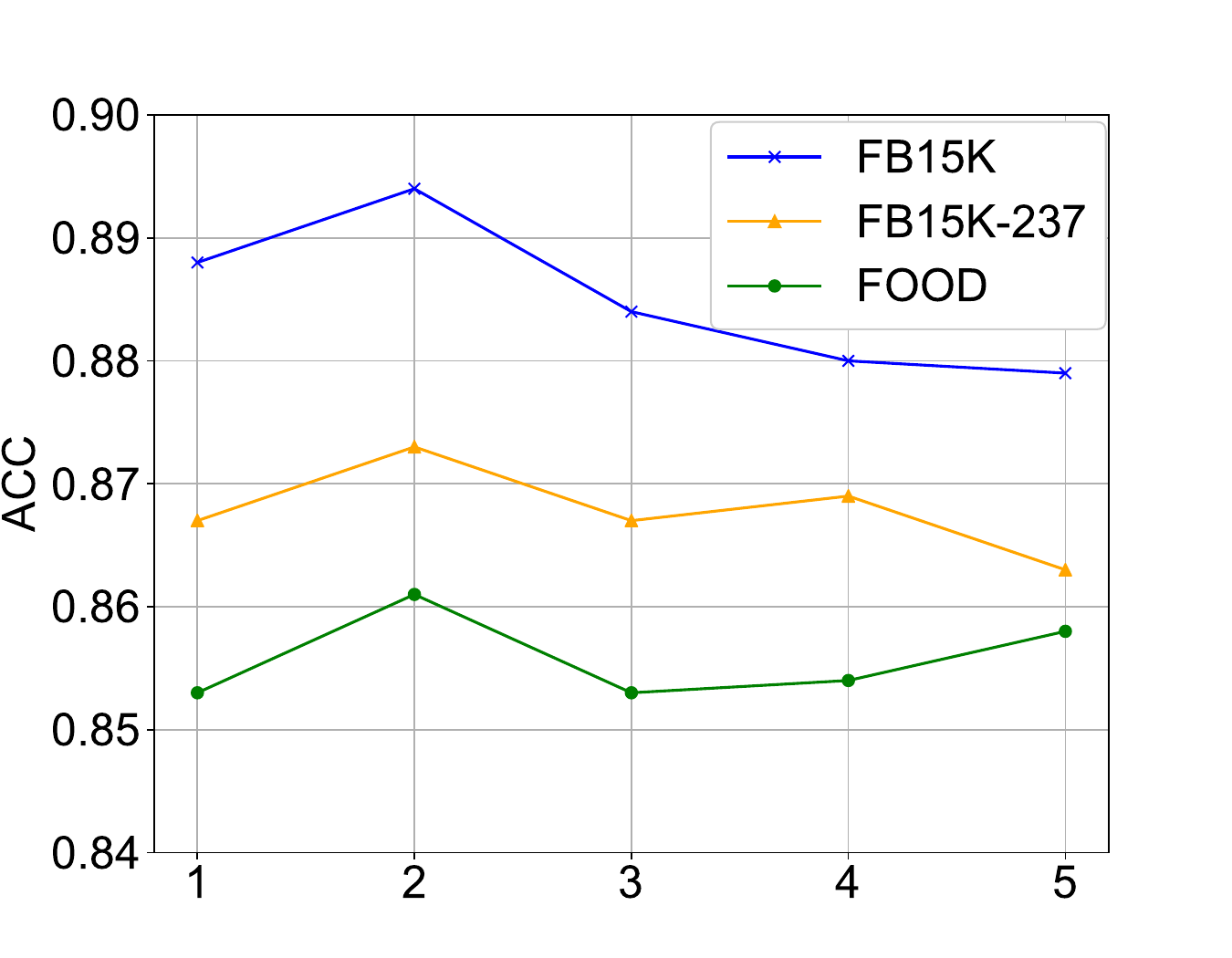}}}
\subfigure[top-$k$ selection]{ 
\includegraphics[width=0.43\textwidth]{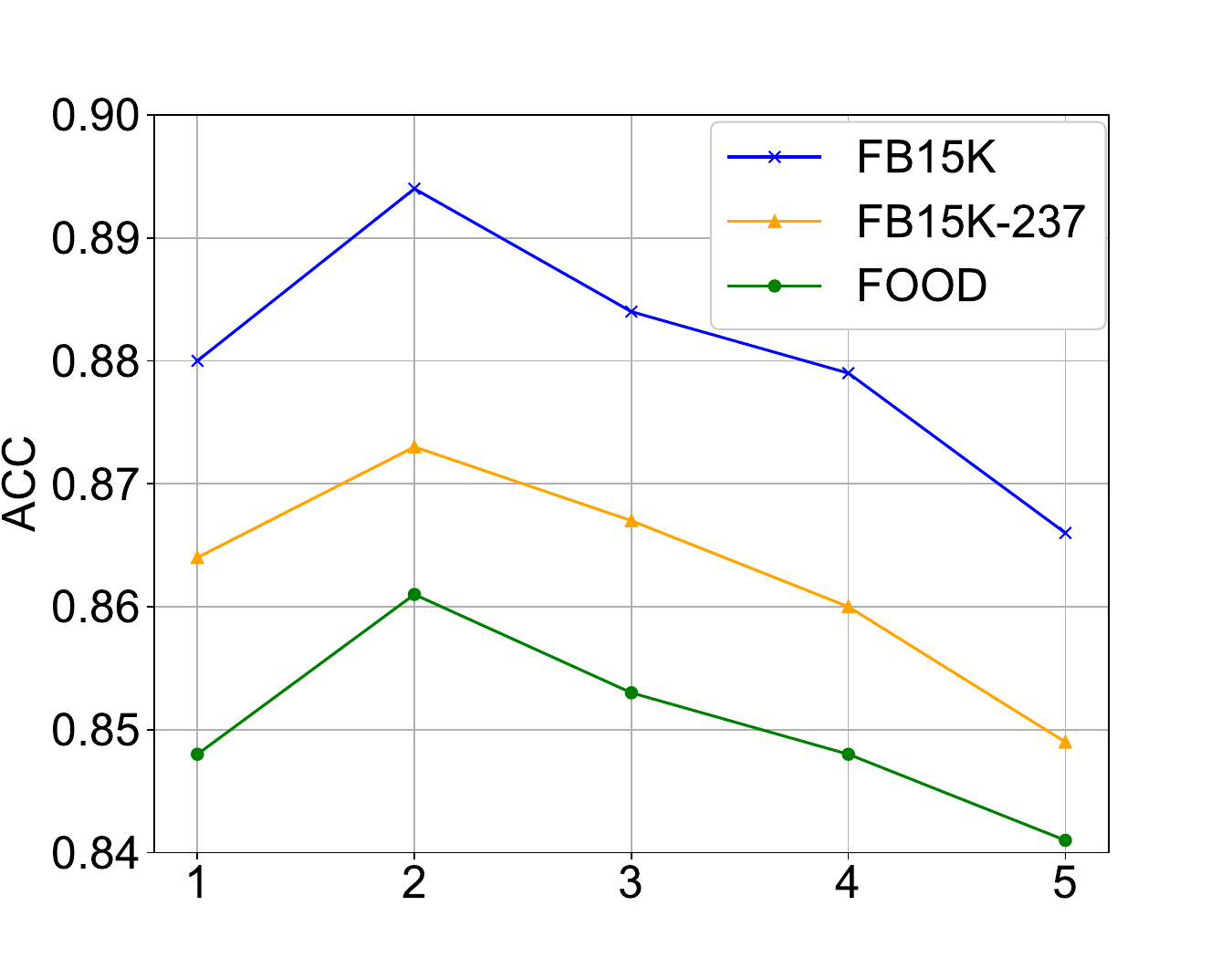}}
\caption{Further analyses of semantic composition module} 
\label{multihead} 
\end{flushleft}
\end{figure}

We then analyze the number of attention heads $n_a$ and top-$k$ selection in the semantic composition module.
Figure~\ref{multihead} shows the results. The performance of multi-head attention is slightly better than that of single-head attention, and using two attention heads is the best. 
The top-$k$ selection represents the semantic composition of $k$ claims according to the attention scores. 
It can be seen from the figure that selecting top-2 claims achieves the best performance, 
indicating that messages passing among 2 claims is the most efficient.
The result demonstrates that our node selection technique is beneficial for semantic composition.

\subsubsection{\textbf{Case Study}}
Table~\ref{casestudy} illustrates two \textit{false} cases of multi-triple claim.
In the table, the claim scores are obtained by Dual TransE, which predicts a plausible score for each triple. The statement scores are obtained by our method.
For both cases, LESC makes the correct prediction while Dual TransE fails.
For case 1, the claim (\textit{tea, lead to, kidney stone}) is false while it is mispredicted by Dual TransE. Our method can learn better representation for the entity \textit{tea} and \textit{kidney stone} that helps to check for the error.
For case 2, the statement conveys a piece of false information: \textit{cherry can cause poisoning}, which cannot be captured by Dual TransE. Our LESC models the interactions among claims, so that it helps to check the truthfulness of the statement accurately.

\begin{table}
\caption{Examples of two \textit{false} multi-claim statements. We predict a score for each triple claim and statement, and the threshold is 0.5.}
\resizebox{1\textwidth}{!}{
\setlength{\abovecaptionskip}{3pt}%
\setlength{\belowcaptionskip}{0pt}%
\centering
  \label{tab:freq}
  \begin{tabular}{c|c|c|c|c}
    \toprule
\textbf{Case}&\textbf{Statement}&\textbf{Triple claim}&\makecell[c]{\textbf{Claim} \\ \textbf{Score}}& \makecell[c]{\textbf{Statement}\\ \textbf{Score}}\\
\midrule
\multirow{3}*{1}&\multirow{3}*{\makecell[c]{Tea contains calcium oxalate, which is one \\of the causes of kidney stones. So drinking \\tea regularly can easily lead to kidney stones.}}&(tea, contain, calcium\_oxalate) &0.84 {\color{Green}\cmark}&\multirow{3}*{0.42 {\color{red}\xmark}}\\
{}&{}&(calcium\_oxalate, cause, kidney\_stone) & 0.70 {\color{Green}\cmark}&{}\\

{}&{}&\cellcolor{mygray}(tea, lead\_to, kidney\_stone) & \cellcolor{mygray}0.53 {\color{Green}\cmark}&{}  \\
    \midrule
\multirow{4}*{2}&\multirow{4}*{\makecell[c]{Cherries are rich in antioxidants, which \\ may reduce chronic disease risk. However,\\ cherries also contain cyanogenic glycoside,\\ which will cause poisoning.}}&(cherry, rich\_in, antioxidants) & 0.79 {\color{Green}\cmark}&\multirow{4}*{0.45 {\color{red}\xmark}}\\
{}&{}&(antioxidants, reduce, chronic\_disease) & 0.58 {\color{Green}\cmark}&{}\\
{}&{}&\cellcolor{mygray}(cherry, contain, cyanogenic\_glycoside) & \cellcolor{mygray}0.82 {\color{Green}\cmark}&{}\\
{}&{}&\cellcolor{mygray}(cyanogenic\_glycoside, cause, poisoning) & \cellcolor{mygray}0.54 {\color{Green}\cmark}&{}\\
  \bottomrule
\end{tabular}
}
\label{casestudy}
\end{table}

\section{Conclusion}
This paper first considers the multi-claim fact checking over a knowledge graph, and proposes an end-to-end learning enhancement and semantic composition model to tackle this problem. 
We propose a KG-based learning enhancement method to learn context-specific representations of entities by selectively aggregating neighboring attributes based on the contextual information.
We then propose a graph-based semantic composition method for verification to effectively compose global and local semantics by devising an enhanced multi-head attention mechanism. 
We construct a real-world dataset on the food domain and then conduct experimental studies on the constructed dataset and two benchmark datasets to validate our LESC. Experimental results demonstrate the effectiveness of our method for multi-claim fact checking.

\section*{Acknowledgments}
This work was supported in part by National Natural Science Foundation of China under Grants \#71621002, \#11832001 and \#71702181.

\section*{References}

\begingroup
\setstretch{0.8}
\bibliography{mybibfile2}
\endgroup
\end{document}